\documentclass[11pt]{article}
\usepackage{coling2020}
\usepackage{times}
\usepackage{url}
\usepackage{latexsym}
\usepackage{color}
\usepackage{siunitx}
\usepackage{listliketab}
\usepackage{multirow}
\usepackage{here}
\usepackage{booktabs}
\usepackage{pdfpages}
\usepackage{framed}
\usepackage{amsmath}
\title{Neural text normalization leveraging similarities of strings and sounds}

\author{\\ \textbf{Riku Kawamura}$^\dagger$\ \ \ \textbf{Tatsuya Aoki}$^\dagger$\ \ \ \textbf{Hidetaka Kamigaito}$^\dagger$\ \ \ \\ \textbf{Hiroya Takamura}$^\dagger$ $^\ddagger$\ \ \ \textbf{Manabu Okumura}$^\dagger$\\ \\
         $^\dagger$Tokyo Institute of Technology\\
         $^\ddagger$National Institute of Advanced Industrial Science and Technology \\ \\
         \{riku, aoki, kamigaito\}@lr.pi.titech.ac.jp \\
          \{takamura, oku\}@pi.titech.ac.jp
         }  

\date{}

\begin{document}
\maketitle
\begin{abstract}
We propose neural models that can normalize text by considering the similarities of word strings and sounds. We experimentally compared a model that considers the similarities of both word strings and sounds, a model that considers only the similarity of word strings or of sounds, and a model without the similarities as a baseline. Results showed that leveraging the word string similarity succeeded in dealing with misspellings and abbreviations, and taking into account the sound similarity succeeded in dealing with phonetic substitutions and emphasized characters. So that the proposed models achieved higher F$_1$ scores than the baseline.
\end{abstract}

\section{Introduction}
\label{sec:intro}
Non-standard words such as misspellings or abbreviations are often used in social media such as Twitter. 
It is difficult to understand the meaning of such words without any extra knowledge, and it is challenging to perform natural language processing over sentences including them.
Text normalization plays an important role in dealing with these broken texts by correcting such a sentence into a standard one. The following is an example of text normalization.

\begin{center}
\textbf{before normalization:}$\,$$\, ${\it \textcolor{red}{r}$\ $$\ $$\,$ \textcolor{red}{u} $\ $$\ $\textcolor{red}{cuming} $\,$\textcolor{red}{2} $\ $midcorner \textcolor{red}{dis} $\ $sunday?}\\
$\ \ \ \ \ \ \ \ \ \ \ \ \ \ \ \ \ \ \ \ \ \ \downarrow\ \ \ \, \downarrow\ \ \ \ \ \ \ \downarrow\ \ \ \ \ \ \ \downarrow\ \ \ \ \ \ \ \ \ \ \ \ \ \ \ \ \ \ \ \ \,\downarrow$\\
 \textbf{after normalization:}$\,\,${\it {\color[rgb]{0,0.7,0.3}{are}} {\color[rgb]{0,0.7,0.3}{you}} {\color[rgb]{0,0.7,0.3}{coming}} {\color[rgb]{0,0.7,0.3}{to}} midcorner {\color[rgb]{0,0.7,0.3}{this}} sunday?}
\end{center}
In text normalization, as shown in the example, we need to correct several types, including misspellings (`{\it cuming}' to `{\it coming}'), abbreviations (`{\it convo}' to `{\it conversation}'), phonetic substitutions (`{\it dis}' to `{\it this}' or `{\it r u}' to `{\it are you}'), and emphasized characters (`{\it yeeeees}' to `{\it yes}').\par 
Even though the similarities of character surfaces and phonemes is important for a human to understand non-standard words, current neural network-based text normalization methods do not consider this information.
We assume that text normalization more intuitive to humans is possible by explicitly considering such features in a neural network-based method.  
Based on this assumption, in this work, we propose neural text normalization models that leverage both string and sound similarities. Experimental results show that our proposed models outperformed a baseline and achieved state-of-the-art results in the text normalization track on WNUT-2015. 

\blfootnote{
    %
    %
    %
    \hspace{-0.65cm}  
     This work is licensed under a Creative Commons 
     Attribution 4.0 International License.
     License details:
     \url{http://creativecommons.org/licenses/by/4.0/}.
    %
    %
}

\section{Related Work}
\label{sec:related}
\newcite{han-etal-2012-automatically} proposed a ranking-based text normalization method that incorporates the matching degree of surrounding word n-grams to the target word and the edit distance from existing words.
\newcite{inproceedings} proposed a text normalization method leveraging phonetic information to translate non-standard words into standard ones. \newcite{Ansari2017ImprovingTN} proposed an automatic optimization-based nearest neighbor matching approach leveraging string and phonetic similarity. \newcite{jin-2015-ncsu} achieved the best performance on the WNUT-2015 task, with a method that generates candidates based on the training data. \newcite{DBLP:journals/corr/abs-1710-03476} extended this work by leveraging additional resources of Twitter and Wikipedia data. However, their method does not take into consideration contextual information. To solve this problem, Sequence-to-Sequence (Seq2Seq) \cite{sutskever2014sequence}  has been used for text normalization. \newcite{DBLP:journals/corr/abs-1904-06100} performed highly accurate and fluent text normalization by using the Attention-based Seq2Seq \cite{bahdanau2014neural}. \newcite{Mani_2020} used Seq2Seq in automatic speech recognition error correction, a task similar to text normalization. Taking these trends into account, in this work, we propose a neural text normalization model that leverages both string and sound similarity.\par




\section{Methodology}
\label{sec:proposed}
\begin{figure*}[t]
\begin{centering}
\includegraphics[width=\textwidth]{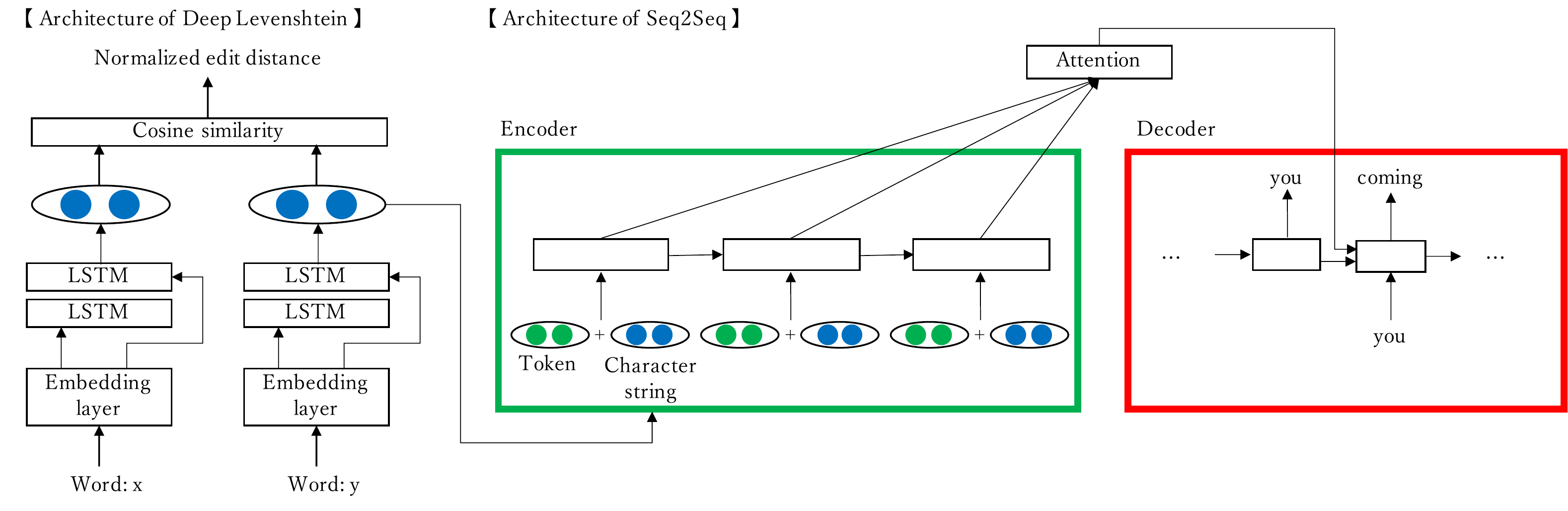}
\vspace{-0.3cm}
\caption{How to introduce the character string feature into Seq2Seq.}
\label{fig:concat}
\end{centering}
\end{figure*}

Figure \ref{fig:concat} shows an overview of our proposed method.
In this study, we perform text normalization based on the method of \newcite{DBLP:journals/corr/abs-1904-06100}, which incorporates token embeddings as an input to the encoder. Our method expands their work by utilizing features related to character strings and sounds.

\subsection{Deep Levenshtein}
\label{sec:deeplevenshtein}
\newcite{moon-etal-2018-multimodal} proposed Deep Levenshtein, a network that captures the feature of character strings based on the Levenshtein edit distance \cite{levenshtein1966bcc} in order to correct any character fluctuations of named entities in a text.
In this study, we incorporate this mechanism into the Seq2Seq model to make it more robust to a broken text.

Deep Levenshtein is a neural network that takes two words $x$ and $y$ as an input and then outputs hidden representations for the character strings of the words.
Two words $x$ and $y$ are fed into the word embedding layer, and then we obtain word embeddings $e_x$ and $e_{y}$, respectively.
Each word embedding is an input to the bidirectional LSTM \cite{Hochreiter1997LongSM} to obtain hidden representations $c_x$ and $c_{y}$, that capture the feature of the character string. Here, $\overrightarrow{h_x}$ and $\overleftarrow{h_x}$ represent the forward and backward paths of the bidirectional LSTM, respectively. $c_x$ is obtained by concatenating them as
$c_x = [\overrightarrow{h_x}; \overleftarrow{h_x}]$,
where ; indicates a concatenation operation.
Deep Levenshtein learns representations $c_x$ and $c_y$ so that the cosine-based similarity between them approximates the similarity based on the edit distance between the input character strings by minimizing

\begin{equation}
    \left|\left|\frac{1}{2}\left(\frac{c_x^{\mathrm{T}} c_{y}}{||c_x|| \cdot ||c_{y}||}+1\right)- s(x,y)\right|\right|^2,
\end{equation}
where $s(x, y)=1-\frac{d(x,y)}{max(length(x), length(y))}$,\footnote{We divide the edit distance, $d(x,y)$, by the longer length of the strings x and y to take into account the length of the character strings.} that indicates the similarity based on the edit distance between the input words $x$ and $y$.

Deep Levenshtein can predict the similarity of the character string between two words based on the distance in vector space, so the vectors obtained above capture the character string feature.

\subsection{Deep Metaphone}
\label{deepmetaphone}
\newcite{raghuvanshi-etal-2019-entity} revealed that relying only on surface text similarities cannot capture phonetic differences between words. Furthermore, \newcite{Han2013LexicalNF} showed that sound-related features are effective in text normalization. In this work, we propose Deep Metaphone to capture sound features for text normalization by learning phonetic edit distance.

Deep Metaphone has the same network structure as Deep Levenshtein.
The difference between them is the training data they use. 
Deep Metaphone learns the phonetic edit distance, which is the edit distance of the strings obtained from Double Metaphone \cite{double-metaphone}. The following is an example conversion where we apply the Double Metaphone algorithm over the words `{\it yeeeees}' and `{\it yes}'.
\begin{center}
$\underset{Before}{\it yeeeees}$ $\rightarrow$ $\underset{After}
{\it AS}$ $\ $ \quad  $\underset{Before}{\it yes}$ $\rightarrow{}$ $\underset{After}{\it AS}$\\
\end{center}
By using Double Metaphone, similar to Deep Levenshtein, Deep Metaphone can predict the sound similarity between two words.
Therefore, we can use the vector that captures the feature of the sound.

\subsection{Incorporating new features to Seq2Seq}
In this study, we use a bi-directional LSTM for the encoder and decoder.
\newcite{DBLP:journals/corr/abs-1904-06100} incorporated only token embeddings $e^{token}_{x_t}$ as an input to the encoder.
We further incorporate the character feature $c^{leven}_{x_t}$, obtained from Deep Levenshtein, and the sound feature $c^{phone}_{x_t}$, obtained from Deep Metaphone.\footnote{Note that Deep Levenshtein and Deep Metaphone take two words as an input during the training steps, but both methods take only a word to extract each feature for Seq2Seq in the inference steps.} 
\newcite{mansfield-etal-2019-neural} empirically tested addition, concatenation, and multi-layer perceptron to combine new features with token embeddings. They reported that concatenation outperforms the other two methods. Therefore, we choose concatenation.
Our new feature vectors $c^{leven}_{x_t}$ and $c^{phone}_{x_{t}}$ are incorporated into the Seq2Seq encoder as follows:
\begin{align}
    &h^{enc}_t = Encoder([e^{token}_{x_t}; c^{leven}_{x_t}; c^{phone}_{x_t}; h^{enc}_{t-1}]).
\end{align}

\section{Experiments}
\label{sec:exp}
{\renewcommand\arraystretch{1.5}
   \begin{table*}[t]
    \begin{center}
      \small
      \begin{tabular*}{160mm}{l|l|l}
      \hline 
      \multicolumn{1}{c|}{Pattern} & \multicolumn{1}{c|}{Description} & \multicolumn{1}{c}{Example} \\ \hline
      1. Do nothing & do not add noise & {\it python} $\rightarrow$ {\it python} \\
      2. Delete characters & delete characters randomly & {\it python} $\rightarrow$ {\it pyhon} \\
      3. Replace characters & replace characters randomly & python $\rightarrow$ {\it pyhtno} \\
      4. Extend characters & extend words ending with \{u, y, s, r\} & {\it beer} $\rightarrow$ {\it beerrrr} \\
      5. Extend short vowels & stretch short vowels \{a, i, u, e, o\} & {\it cat} $\rightarrow$ {\it caaat} \\
      6. Delete symbol & delete apostrophe & {\it I'm} $\rightarrow$ {\it Im} \\
      7. Misplaced sign & insert apostrophe in different position & {\it don't} $\rightarrow$ {\it do'nt} \\
      8. Typo & replace with another character that is at a near position on a keyboard & {\it hello} $\rightarrow$ {\it jello} \\
      9. Convert to another token & convert to completely different token & {\it python} $\rightarrow $ {\it ruby} \\
       \hline
      \end{tabular*}
      \end{center}
      \label{tab:noise}
      \caption{Noise generator.}
\end{table*}
}

\subsection{Experimental settings}
We used WNUT-2015 Shared Task2 \cite{baldwin-etal-2015-shared},\footnote{https://noisy-text.github.io/2015/} which is a task of normalizing social media texts, for our evaluation.
The official dataset consists of 4,917 tweets with 373 non-standard words. The dataset was randomly split by \newcite{DBLP:journals/corr/abs-1904-06100} into 60:40, 2,950 tweets for the training data and 1,967 tweets for the test data.
For evaluation metrics, we used precision, recall, and F$_1$-score. 

To train Deep Levenshtein, we applied the noise generator in \cite{DBLP:journals/corr/abs-1904-06100} to words in the training data of WNUT-2015 Shared Task2. 
The noise generator outputs character strings by executing one of the processes in Table 1.
We used both those generated words and the original words to train Deep Levenshtein.

For training Deep Metaphone, we first created the training data for Deep Levenshtein as in the above and then applied the Double Metaphone algorithm to words to yield the training data.

Parameters including the size of the token embeddings in the baseline model were set to the same as in \cite{DBLP:journals/corr/abs-1904-06100}, where it was 100. 
The sizes of hidden layers, $c^{leven}_{x_t}$ and $c^{phone}_{x_t}$, of LSTM for Deep Levenshtein and Deep Metaphone were tuned from \{10, 20, 30, 40, 50\} on the validation data, randomly extracted 100 sentences from the training data. When only Deep Levenshtein was used, 20 was selected. When only Deep Metaphone was used, 50 was selected. When both were used, 
10 was selected for each of them.
The reason why we set the range of smaller values from 10 to 50 compared with the size of token embeddings, 100, is based on the finding in \cite{sennrich-haddow-2016-linguistic}. They reported that the size of the secondary embeddings, i.e., $c^{leven}_{x_t}$ and $c^{phone}_{x_t}$ in our case, should be smaller than that of the primary embeddings, i.e., $e^{token}_{x_t}$ in our case.

\subsection{Compared models}
In the experiments, we compared a baseline model and our models, which are listed below.
\vspace{1.5mm}\\
    {\bf Two-stage Seq2Seq (baseline)}: A model proposed by \newcite{DBLP:journals/corr/abs-1904-06100}. We reimplemented this model and will report its performance in addition to the scores reported in their paper.
    \vspace{1.5mm}\\
    {\bf Two-stage Seq2Seq + LS (Levenshtein)}: A model where we add character string features to the input of the encoder in the Two-stage Seq2Seq model. \vspace{1.5mm} \\
    {\bf Two-stage Seq2Seq + MP (Metaphone)}: A model where we add sound features to the input of the encoder in the Two-stage Seq2Seq model. \vspace{1.5mm} \\
    {\bf Two-stage Seq2Seq + LS + MP}: A model where we add both character string and sound features to the input of the encoder in the Two-stage Seq2Seq model.

{\renewcommand\arraystretch{1.6}
\begin{table*}[t]
    \begin{center}
      \small
      \begin{tabular*}{102.5mm}{l|ccc}
      \hline
      Model & Precision & Recall & F$_1$ \\ \hline
      Two-stage Seq2Seq (reported) & 90.66 & 78.14 & 83.94 \\ 
      Two-stage Seq2Seq (reproduced)  & 90.24 & 78.10 & 83.74 \\
      Two-stage Seq2Seq + LS & \textbf{91.89} & 78.00 & \textbf{84.38} \\ 
      Two-stage Seq2Seq + MP & 91.32 & 78.18 & 84.24 \\ 
      Two-stage Seq2Seq + LS + MP & 91.30 & 77.80 & 84.13 
      \\
      Random Forest \cite{jin-2015-ncsu} & 90.61 & \textbf{78.65} & 84.21
      \\
      \hline
      MoNoise \cite{DBLP:journals/corr/abs-1710-03476}\footnotemark[5] & 93.53 & 80.26 & 86.39 \\
      \hline
      \end{tabular*}
      \label{result:eval}
    \end{center}
     \caption{Comparison of our models with the baseline and state-of-the-art model on WNUT-2015.\footnotemark[4]
     \textbf{Bold} indicates the best score.
     For reference, we also list the result of MoNoise \cite{DBLP:journals/corr/abs-1710-03476}, which incorporates data augmentation with external resources.\footnotemark[
    5]}
\end{table*}
\footnotetext[4]{While the performance degraded when the token embedding size was set to a larger value, 120, it was improved when combining the character string/sound features.}
\footnotetext[5]{In addition to the WNUT-2015 training data, MoNoise leverages large collections of Twitter and Wikipedia data.}
}

{\renewcommand\arraystretch{1.5}
\begin{table}[t]
    \begin{center}
      \small
      \begin{tabular*}{160mm}{l|l|l|l|l}
      \hline
      Type & Misspelling & Abbreviation & Phonetic substitution & Emphasized character\\ \hline
      Input  & {\it wen u cee a car} & {\it i diss you} & {\it so funny d temple} & {\it rt: i go homeee}\\ \hline
      Reference & {\it when you see a car} & {\it i disrespect you} & {\it so funny the temple} & {\it rt: i go home} \\ \hline
      Two-stage Seq2Seq & {\it \textcolor{red}{wen} you see a car} & {\it i \textcolor{red}{diss} you} & {\it so funny \textcolor{red}{d} temple} & {\it rt: i go \textcolor{red}{homeee}}\\ \hline
      Two-stage Seq2Seq + LS & {\it {\color[rgb]{0,0.7,0.3}{when}} you see a car} & {\it i {\color[rgb]{0,0.7,0.3}{disrespect}} you} & {\it so funny \textcolor{red}{d} temple} & {\it rt: i go {\color[rgb]{0,0.7,0.3}{home}}}\\ \hline
      Two-stage Seq2Seq + MP & {\it \textcolor{red}{wen} you see a car} & {\it i {\color[rgb]{0,0.7,0.3}{disrespect}} you} & {\it so funny {\color[rgb]{0,0.7,0.3}{the}} temple} & {\it rt: i go {\color[rgb]{0,0.7,0.3}{home}}}\\ \hline
      Two-stage Seq2Seq + LS +MP & {\it {\color[rgb]{0,0.7,0.3}{when}} you see a car} & {\it i \textcolor{red}{this} you} & {\it so funny \textcolor{red}{d} temple} & {\it rt: i go {\color[rgb]{0,0.7,0.3}{home}}} \\ \hline
      \end{tabular*}
    \end{center}
    \caption{Examples for correcting several types.}
    \label{tab:variants}
\end{table}
}  

\subsection{Results and analysis}
We report the performance of the baseline model and our models in Table 2.
As shown, our models, i.e., Two-stage Seq2Seq +LS, +MP, +LS +MP, outperformed the baseline model in terms of F$_1$ score.
Table~\ref{tab:variants} shows example outputs from the Two-stage Seq2Seq model and our models.
From the table,
we can see that Two-stage Seq2Seq + LS and + LS + MP models corrected the misspelled `{\it wen}' to `{\it when}'. Two-stage Seq2Seq + LS and + MP models corrected the abbreviation `{\it diss}' to `{\it disrespect}'.
Only Two-stage Seq2Seq + MP model corrected the phonetic substitution `{\it d}' to `{\it the}'. 
All of our models corrected the emphasized character `{\it homeee}' to `{\it home}'.
These results show that Two-stage Seq2Seq + LS model works well for misspellings and abbreviations and that Two-stage Seq2Seq + MP model works well for phonetic substitutions and emphasized characters.

On the contrary, the weakness of Two-stage Seq2Seq + LS model is that it fails to deal with typos where we need to correct from a character to another character that is far on the keyboard layout, such as `{\it thang}' to `{\it thing}'. 
The weakness of Two-stage Seq2Seq + MP model is that it tends to mistakenly correct a word to another word having a similar sound, such as `{\it nah}' to `{\it no}'. 
We also found that, when considering both features, we observed several error cases such as `{\it favor}' to `{\it favorite}', where the model wrongly changed a word. We speculate that these are due to the small search range of the sizes of the hidden layers ($c^{leven}_{x_t}$ and $c^{phone}_{x_t}$).
Furthermore, all models were not able to handle the correction of an emoji, such as `{\it b}', meaning thumbs-up.
This is because the knowledge in the current models is based only on the training data, and we do not utilize specific knowledge on the connection between linguistic information and other types of information, including visual information, which we leave for our future work.




\section{Conclusion and future work}
\label{sec:conclusion}

In this paper, we proposed a method that takes into account the similarities of word strings and sounds as features for text normalization.
Our evaluation results showed that incorporating such features to Seq2Seq improves the performance of text normalization compared to the baseline method in terms of F$_1$ score. We also found that the proposed methods managed to effectively consider both surface character and phonetic similarities.\par
For future work, we will try data augmentation based on the finding by \newcite{grundkiewicz-etal-2019-neural}. They reported that synthetic data can be generated by substituting words commonly confused with each other to improve performance. On the contrary, we will use the dataset proposed by \newcite{DBLP:journals/corr/SproatJ16}, a much larger dataset than the WNUT-2015 Shared Task2, to empirically gain deep insights. We would also like to apply ELMo \cite{Peters_2018}, Transformer \cite{NIPS2017_7181} and BERT \cite{Devlin_2019} to Seq2Seq for better performance.

\bibliographystyle{acl}
\bibliography{ref}

\end{document}